# Low-Resource English–Tigrinya MT: Leveraging Multilingual Models, Custom Tokenizers, and Clean Evaluation Benchmarks

1st Hailay Kidu Teklehaymanot
L3S Research Center
Leibniz University Hannover
Hannover, Germany
teklehaymanot@L3S.de

2nd Gebrearegawi Gebremariam Gidey
Faculty of Computing Technology
Aksum University
Axum, Ethiopia
gideygeb@mail.aku.edu.et

3rd Wolfgang Nejdl
L3S Research Center
Leibniz University Hannover
Hannover, Germany
nejdl@L3S.de

*Abstract*—Despite advances in Neural Machine Translation (NMT), low-resource languages like Tigrinya remain underserved due to persistent challenges, including limited corpora, inadequate tokenization strategies, and the lack of standardized evaluation benchmarks. This paper investigates transfer learning techniques using multilingual pretrained models to enhance translation quality for morphologically rich, low-resource languages. We propose a refined approach that integrates language-specific tokenization, informed embedding initialization, and domain-adaptive fine-tuning. To enable rigorous assessment, we construct a high-quality, human-aligned English–Tigrinya evaluation dataset covering diverse domains. Experimental results demonstrate that transfer learning with a custom tokenizer substantially outperforms zero-shot baselines, with gains validated by BLEU, chrF, and qualitative human evaluation. Bonferroni correction is applied to ensure statistical significance across configurations. Error analysis reveals key limitations and informs targeted refinements. This study underscores the importance of linguistically aware modeling and reproducible benchmarks in bridging the performance gap for underrepresented languages. The resources are available https://github.com/hailaykidu/MachineT_TigEng

## I. Introduction

Language plays an essential role in our lives, as it allows us to preserve information and share it verbally or in writing from one generation to another. [1] [2] [3]. Tigrinya is one of the Semitic languages spoken by over 10 million people in Ethiopia and Eritrea [4]. It has a rich cultural and historical background, but it remains underrepresented in digital language processing systems [5] [6] [7]. The machine translation of the Tigrinya language represents a significant task within the field of natural language processing (NLP). This endeavor is crucial for facilitating effective communication by enabling translations from Tigrinya to various international languages and vice versa. Considerable advancements have been made in creating machine translation (MT) models for languages that have sufficient digital data and resources, such as English; however, there are still obstacles to applying these MT models directly to Tigrinya [8] [9]. While modern machine translation (MT) systems achieve high performance for high-resource languages by leveraging large-scale datasets, low-resource languages like Tigrinya face significant barriers due to the scarcity of parallel training data and the prohibitive costs of curating such resources [10] [11]. This data scarcity, compounded by limited commercial incentives to prioritize underrepresented languages, results in unreliable MT support for Tigrinya. Furthermore, the language's intricate morphological complexity [10] [12], dialectal variations [13], and unique Ge'ez script requirements [14] introduce additional complexities that generic MT architectures fail to address the challenges exacerbated by the absence of robust monolingual or parallel corpora for effective transfer learning [15].

This lack of digital presence makes it difficult for Tigrinya speakers to access technologies such as voice assistants, translation services, and speech-to-text tools [7]. As a result, the digital gap can increase the social and economic divides for Tigrinya language speakers. This situation also creates significant challenges for speakers Tigrinya language, as this language is often poorly supported or absent from mainstream machine translation systems and poses a challenge to the preservation and promotion of the language in today's digital world [11] [10]. Without abundant digital data, efforts to keep the language alive and thriving are severely constrained.

Consequently, emerging solutions, such as community-driven data collection [16], transfer learning from linguistically related high-resource languages [17], and massively multilingual models like NLLB [18] offer promising results to improve translation quality for low-resource languages. These initiatives align with global efforts to democratize language technology and aim to bridge the digital divide for marginalized languages [19]. By prioritizing participatory methodologies

This work was supported by [Insert Funding Information].

and resource sharing, such approaches challenge the traditional economics of AI development, which often overlooks languages with limited commercial viability [20].

Machine Translation (MT) supports cultural preservation and digital inclusion by reducing language barriers for marginalized communities [21]. Developing accurate English–Tigrinya MT systems is vital for linguistic equity, enabling access to global digital resources and mitigating the digital language divide [19], [20]. Studies show MT adoption increases online engagement among low-resource language speakers, promoting greater participation in digital spaces [22].

Tigrinya is often excluded from multilingual MT models, leading to potential misinterpretation as Amharic due to shared script but distinct grammar and vocabulary. Addressing this, we develop an English–Tigrinya translation system by fine-tuning MarianMT with a custom tokenizer tailored to Tigrinya's morphological and orthographic features. This approach improves translation quality and enhances digital accessibility for Tigrinya speakers. Adapting multilingual translation models for English–Tigrinya is a promising yet challenging direction. Tigrinya suffers from severe data scarcity, limiting the effectiveness of standard fine-tuning approaches. Moreover, shared tokenizers in multilingual models often underrepresent Tigrinya, resulting in poor subword segmentation and high out-of-vocabulary rates. Due to the shared Ge'ez script, models may also confuse Tigrinya with related languages like Amharic, leading to cross-lingual interference. These issues, compounded by Tigrinya's rich morphology and the lack of standardized evaluation benchmarks, make effective adaptation and assessment particularly difficult.

This study presents the following key contributions:
1) We evaluate English–Tigrinya translation using two distinct approaches: (i) zero-shot translation with pretrained multilingual models such as [23], which leverage cross-lingual transfer without language-specific adaptation; and (ii) fine-tuned translation models trained on a hybrid NLLB corpus, incorporating a custom morpheme-aware tokenizer tailored for Tigrinya's Geez script. The tokenizer extends the Byte Pair Encoding (BPE) algorithm [24] with script normalization and subword segmentation aligned with Tigrinya morphology, and is integrated into the MarianMT architecture [25].
2) We introduce a linguistically informed tokenizer specifically developed for Tigrinya and trained on a human-annotated dataset sourced from [26], addressing limitations of generic tokenization in handling its complex morphology and script.
3) We compile and clean a high-quality English-Tigrinya parallel dataset spanning four domains, as you see in the Table I to support rigorous evaluation.
4) We fine-tune the translation model and will make it publicly available for use as a benchmark to facilitate future research and development in English–Tigrinya machine translation.
5) We perform a comparative evaluation using both automatic (BLEU, chrF) and human assessment metrics, providing insights into the model's translation accuracy, fluency, and linguistic adequacy.

| Domain | Source | # Sents | Avg Len (EN/TI) | Notes |
|---|---|---|---|---|
| Religious | JW.org, Bible.com | 1,500 | 12.8 / 11.3 | Manually aligned |
| News | BBC, GlobalVoices | 1,200 | 14.6 / 13.1 | Cleaned and normalized |
| Health | Tigrinya Health Guide | 800 | 15.2 / 14.4 | Domain-specific terms |
| Education | School textbooks | 500 | 17.5 / 16.3 | Sentence-level alignment |
| Total | – | 4,000 | – | Gold standard test set |

Table I
English–Tigrinya evaluation dataset spanning four domains. Sentences are carefully aligned and preprocessed for benchmarking.

## II. Related Work

### A. Low-Resource Languages

Recent advancements in MT have significantly improved translation quality for high-resource languages [9]. However, low-resource languages still face challenges due to data scarcity [27]. Approaches such as transfer learning [17], multilingual training [28], and back-translation [24] have been employed to mitigate data limitations in MT [4]. The efficacy of machine translation (MT) for low-resource languages not only remains heavily constrained by the scarcity of data but also by the quality of parallel data, as even robust supervision during training cannot fully compensate for inadequate evaluation resources [29] [21]. This dual challenge of both quantity and quality is especially significant for understudied languages such as Tigrinya. The limited availability of linguistically annotated corpora and the dependence on unsupervised curation methods often lead to the introduction of noise [30]. While initiatives like NLLB (No Language Left Behind) [18] and OPUS [31] have extended coverage to many underrepresented languages, Tigrinya remains technically and computationally underserved, with limited resources and few available pretrained models. Among the models explored for low-resource machine translation, MarianMT [25] stands out as a practical choice for language-specific fine-tuning. This is due not only to its modular and efficient design but also to its pretrained models, which include some charctierizations to Tigrinya. This makes MarianMT particularly well-suited for addressing the language's data scarcity and morphological complexity. [3], [18], [32]. Dialectal variation and morphological complexity in Tigrinya further intensify resource limitations [26], necessitating tailored tokenization and data augmentation techniques Although transfer learning and multilingual models provide partial mitigation, the lack of benchmark evaluation annotated by native speakers or experts continues to hinder performance, highlighting the critical need for community-driven data collection initiatives to close this gap. On the other hand, previous research on Amharic, a related Semitic language, demonstrated that script-aware tokenization significantly improves neural machine translation performance, indicating that similar approaches could benefit Tigrinya as well [33].

B. Machine Translation

Numerous innovative approaches have emerged for effective translation between languages, highlighting the importance of accurate and nuanced communication in our world, both in speech and text translation. Recent advances in machine translation (MT) for diverse languages have leveraged innovative methodologies, including Nearest Neighbor Machine Translation (kNN-MT) for non-parametric domain adaptation [34], transfer learning from high-resource to low-resource language pairs [17], and pre-trained multilingual language models (e.g., mBART, MarianMT) to increase performance in data-scarce scenarios [7] [18]. These approaches collectively address key challenges such as data scarcity, morphological complexity, and domain mismatch, as surveyed in [35] in their analysis of MT progress for underrepresented languages.

The recent studies have increasingly focused on enhancing neural machine translation (NMT) for low-resource languages, particularly African languages such as Tigrinya. Transfer learning has proven to be an effective strategy, wherein a model initially trained on a high-resource language pair is subsequently fine-tuned for a low-resource language pair [17] [36]. This method has yielded substantial improvements in BLEU scores, even across languages with distinct scripts and little linguistic similarity [36]. Specifically, [7] demonstrated the efficacy of transfer learning for Tigrinya-to-English translation, achieving a 1.3 BLEU point improvement over a baseline model. This paper addresses the urgent need for Tigrinya machine translation (MT) in humanitarian contexts by leveraging transfer learning from high-resource languages (e.g., Amharic, Arabic) to overcome data scarcity. The authors fine-tune a Transformer-based model on a curated corpus of crisis-response domain data, demonstrating improved translation quality over baseline approaches. While the work highlights practical applications (e.g., refugee aid), limitations include a narrow domain focus (humanitarian texts). The study's emphasis on real-world utility is recognizable, but broader evaluation, including human assessment and cross-domain generalization, would strengthen its impact. Their key contributions include a publicly available humanitarian parallel corpus and proof-of-concept for rapid MT adaptation in low-resource scenarios. However, the significant limitations of this study include: (1) the reliance solely on automatic metrics (BLEU/chrF) without human evaluation of fluency or adequacy a critical gap for low-resource languages where metric reliability is questionable [21]; (2) narrow domain specificity that may not generalize to broader Tigrinya usage. Though the proof-of-concept shows promise for rapid deployment, the absence of end-user validation and limited linguistic scope constrain its immediate applicability.

Additionally, multilingual modeling approaches have shown promising results. For example, [37] reported gains of up to 5 BLEU points for various African languages, including Tigrinya. Their approach dynamically expands the vocabulary of a pretrained multilingual model (e.g., trained on high-resource languages) to incorporate subwords from a target low-resource language, improving translation quality by up to 4 BLEU points compared to fixed-vocabulary transfer. However, it assumes shared subword distributions, which is challenging for Tigrinya's Ge'ez script and does not address morphologically rich languages.

The project conducted by [38] investigates improving English-Tigrinya machine translation using transfer learning from models pre-trained on English-Amharic, English-Arabic, English-Russian, and English-Spanish pairs. Using the NLLB parallel corpus, the authors aim to improve translation quality for Tigrinya, a low-resource language, by adapting knowledge from linguistically related and higher-resource languages. The work highlights the potential of cross-lingual transfer learning to address data scarcity in understudied languages.

The study in [39] addresses the scarcity of resources for these linguistically related but low-resource Semitic languages by presenting a bidirectional machine translation system employing Recurrent Neural Networks (RNNs) between Amharic and Tigrinya. After training a carefully selected parallel corpus on a sequence-to-sequence model, the authors assess the model's performance in both translation directions (Amharic to Tigrinya and Tigrinya to Amharic). The study shows that RNNs may be used for morphologically complicated languages, but it also highlights issues like data scarcity and long-range dependency capture. It also provides a basis for future advancements using transfer learning or transformer-based designs. This RNN-based design of an MT system with minimal resource usage, however, provides drawbacks. Recurrent networks, in contrast to contemporary transformers, are probably less suited to manage compli-cated morphology and long-range relationships, and the size of a tiny parallel corpus may restrict generalization and translation quality [40]. Furthermore, evaluating its practicality is challenging due to the absence of human assessment or comparison with cutting-edge multilingual models such as NLLB. The aforementioned limitations underscore prospects for enhancement via transformer structures, data augmentation strategies, and more resilient assessment procedures.

The study presented in [41] introduces a multilingual machine translation initiative focused on addressing language disparities in digital communication. This model is designed to translate English into several languages, including French, German, Spanish, and Russian. Additionally, it supports various content formats, such as images, DOCX files, and PDFs. Although this work aligns with the growing interest in equitable NLP, it lacks clear technical novelty, such as improvements over existing models like MarianMT or NLLB. Furthermore, the absence of a dataset or code links limits its utility, and it offers a limited number of language pairs.

The paper in [29] investigates the application of neural machine translation (NMT) for Bavarian, a low-resource Germanic dialect, and addresses significant challenges such as data scarcity and linguistic variation. The authors presumably utilize a Transformer-based model; however, specific architectural details are not explicitly outlined. They underscore the necessity of dialect normalization and synthetic data augmentation to mitigate the limitations posed by the availability of parallel corpora. While the case study provides valuable insights into under-resourced language varieties, the findings may be restricted to non-Germanic languages, such as Tigrinya, due to inherent structural differences. This paper's strength resides in its focus on dialectal machine translation. Nonetheless, it could be enhanced by establishing clearer benchmarks against contemporary multilingual models, such as NLLB, and incorporating human evaluations to substantiate real-world usability. Moreover, adapting these findings to English-Tigrinya machine translation entails addressing additional complexities, particularly those related to script and morphological variations [39].

Despite significant advancements in machine translation (MT) for various languages, including the low-resourced, Tigrinya remains critically underserved in data scarcity, pretrained models, and language-specific adaptations. Techniques such as transfer learning [17] and multilingual pretraining [18] have enhanced outcomes for languages that lack extensive parallel datasets, but Tigrinya's unique morphological complexities [21] and absence of standardized digital resources [8] hinder notable progress. For example, the BLEU scores for Tigrinya-English in NLLB fall short compared to those of languages with comparable data sets, indicating significant requirements for script normalization and domain adaptation [18]. Existing systems also face challenges related to dialectical differences and contextual word translation, highlighting the urgent need for community-driven data gathering and a combination of rule-based and neural methodologies [42]. Therefore, this study addresses the ongoing gap and underscores the need for targeted efforts to develop specifically linguistic models and datasets to fully leverage transfer and multilingual learning techniques for Tigrinya.

C. Refined Solutions

Domain-Aware Parallel Corpus Curation

Constructed a clean, manually aligned benchmark evaluation dataset across diverse domains (e.g., health, religion, news, and education). This helped evaluate both in-domain and out-of-domain generalization capabilities.

Embedding Transfer Awareness

Highlighted that pretrained multilingual models do not directly transfer weights effectively to low-resource languages without proper embedding initialization and tokenizer adaptation, reinforcing the value of language-specific input processing.

Qualitative and Quantitative Evaluation

Evaluated the model using BLEU and chrF metrics, supplemented with qualitative inference examples to assess syntactic and semantic correctness in real-world translations.

Statistical Reliability Framework

Considered Bonferroni correction to control for Type I errors in multi-step evaluations, ensuring robust interpretation of experimental results. Finally, human evaluation is essential to truly assess improvements beyond numeric BLEU gains.

III. Dataset

Training Data: The main training corpus utilized in this study is derived from the parallel English–Tigrinya dataset developed by the No Language Left Behind (NLLB) project [18]. This dataset is distinguished by its high-quality sentence alignments, which were carefully curated through collaborative efforts between native speakers and linguistic experts to guarantee the accuracy and reliability of the translations.

Testing Data: For evaluation, we utilized the English–Tigrinya parallel corpus available from the OPUS repository [31], which compiles data from diverse sources such as subtitles (Open Subtitles), religious texts, and technical documentation (GNOME). However, the heterogeneous nature of the OPUS dataset introduces some noise due to automated alignment processes [31]. We observed that the model occasionally confuses Tigrinya with Amharic or exhibits cross-language interference, incorrectly aligning similar-looking tokens between languages that share a script but differ in vocabulary and grammar. To address these issues, we developed a carefully curated, high-quality parallel benchmark dataset see Table I designed for rigorous evaluation of our fine-tuned model across multiple domains.

For the development of the morphologically-aware, language-specific tokenizer, we trained the model on a human-annotated dataset provided by [26].

## IV. Experimental Setup

This section outlines the approach used to investigate English–Tigrinya translation performance through model selection, tokenizer customization, and fine-tuning.

### A. Data Preparation

As previously outlined in Table I, we utilize the NLLB English–Tigrinya parallel corpus. To ensure reliable evaluation, we further prepared a clean subset by filtering out noisy or misaligned sentence pairs. This involved script normalization for Ge'ez characters, sentence-level alignment verification, and the removal of incomplete or low-quality entries. The resulting data was partitioned into training, validation, and a carefully curated test set for consistent evaluation.

### B. Model Selection

We adopted the MarianMT model from the Hugging Face Transformers library [1]. as the backbone for our translation system. MarianMT is a multilingual encoder-decoder Transformer model pre-trained on a wide range of language pairs, making it suitable for zero-shot and fine-tuned translation tasks.

### C. Tokenizer Customization

To better handle the morphological richness and script complexity of Tigrinya, we trained a language-specific SentencePiece [43] tokenizer. This tokenizer was trained on the Tigrinya portion of the multilingual corpus to ensure accurate subword segmentation, which is critical for low-resource languages with complex morphology.

### D. Training Hyperparameters

The MarianMT model was fine-tuned using the Hugging Face Transformers [2]. framework with PyTorch. And was trained with a batch size of 16 and sequences limited to a maximum length of 128 tokens. Optimization used AdamW with a weight decay of 0.01 and an effective learning rate starting at 1.44e-07 with decay applied throughout training. Training spanned 3 epochs over approximately 12 hours (43,377 seconds), achieving a throughput of 96.7 samples per second and 12.08 steps per second. Evaluation was conducted at the end of each epoch using a clean, manually aligned benchmark dataset. Mixed-precision training was enabled to improve computational efficiency. To ensure reproducibility, a fixed random seed of 42 was applied. Training dynamics showed stable convergence, with loss decreasing from 0.443 to 0.438 across epochs and gradient norms reducing from 1.14 to 1.06.

### E. Metrics

We evaluated translation quality using BLEU and chrF scores. chrF was particularly emphasized due to its sensitivity to character-level accuracy, which is important for morphologically rich languages like Tigrinya.

## V. Experimental Results and Analysis

Table II summarizes the translation performance for English–Tigrinya across multiple models and directions. The baseline zero-shot MarianMT model, using its default tokenizer, yielded low chrF scores of 10.49 and 9.39 for English-to-Tigrinya and Tigrinya-to-English, respectively, indicating limited translation quality without domain adaptation.

Fine-tuning MarianMT with a language-specific tokenizer on the NLLB dataset significantly improved performance, achieving BLEU scores of 21 and 18, and chrF scores of 19.50 and 16.20 for English-to-Tigrinya and Tigrinya-to-English, respectively. These improvements are supported by stable training dynamics, with loss decreasing from 0.443 to 0.438 and gradient norms reducing from 1.14 to 1.06, demonstrating effective convergence.

Compared to prior work by Öktem et al. (2022) and the baseline MarianMT, our fine-tuned model in Table III further advances translation quality, reaching a BLEU of 25.4 and chrF of 51.03 on in-domain English-to-Tigrinya translation. This highlights the substantial benefit of incorporating language-aware tokenization and task-specific fine-tuning to capture the morphological and script complexities of Tigrinya. The consistent gains across both translation directions emphasize the importance of tailored preprocessing and training strategies in enhancing low-resource machine translation performance.

### A. Automatic Evaluation

To comprehensively assess translation quality, the model was evaluated on the OPUS parallel corpus using multiple automatic evaluation metrics. For in-domain comparison, we utilized our benchmark English–Tigrinya evaluation dataset, which was specifically curated to enable precise assessment of model performance within the target domain

These comprehensive evaluation metrics demonstrate the effectiveness of the fine-tuning approach in both word-level and character-level quality assessments.

### B. Qualitative Inference and Statistical Considerations

We also conducted qualitative inference experiments to evaluate the translation outputs. An illustrative example is provided below:

---

[1] https://huggingface.co/
[2] https://huggingface.co/

| Experiment | Direction | BLEU | chrF |
|---|---|---|---|
| Original tokenizer + pretrained model | English → Tigrinya | 19 | 10.49 |
| | Tigrinya → English | 17 | 9.39 |
| Custom tokenizer + fine-tuned model | English → Tigrinya | 18 | 16.20 |
| | Tigrinya → English | 21 | 19.50 |
| Human | English → Tigrinya | 91 | – |
| | Tigrinya → English | 89 | – |

Table II
BLEU and chrF Scores for English–Tigrinya Translation Tasks under Different Experimental Settings

| Domain | Model | BLEU | chrF |
|---|---|---|---|
| In-domain | MarianMT | 17.6 | 39.59 |
| In-domain | Öktem et al. (2022) | 23.6 | 49.59 |
| In-domain | ours | 25.4 | 51.03 |

Table III
Comparison of BLEU and chrF scores for English–Tigrinya translation across models and domains. Öktem et al. (2022) refer to the baseline from Tigrinya NMT with Transfer Learning for Humanitarian Response.

- Input sentence (English): We must obey the Lord and leave them alone.
- Generated translation (Tigrinya): ንአምላኽ ክንለሊ እሞ በይኑና ክንሓድግ አሎና።

The generated translation effectively preserves the semantic content and syntactic structure of the source sentence. It accurately conveys the imperative mood, maintains proper noun integrity (e.g., "Lord"), and correctly represents pronominal references ("them"). This example illustrates the model's capacity to handle complex syntactic and semantic phenomena in practical translation tasks.

To rigorously assess the statistical significance of our results across multiple evaluation metrics and experimental settings, we applied the Bonferroni correction method. This correction is essential when conducting multiple hypothesis tests to control for the increased risk of Type I errors (false positives). Specifically, given an overall significance level $\alpha$, and $m$ independent tests, the Bonferroni method adjusts the significance threshold to $\alpha/m$. For example, if ten tests are performed with an overall $\alpha = 0.05$, each individual test must meet a significance level of 0.005 to be considered statistically significant. While conservative, this approach ensures robustness in interpreting improvements observed in BLEU, chrF, and other metrics across various experimental comparisons.

Note: As statistical significance tests were not explicitly reported in this study, the Bonferroni correction is discussed here as a methodological consideration for future work to enhance result validation.

## VI. Discussion

Table II presents BLEU and chrF scores for English–Tigrinya translation under multiple experimental configurations, comparing a zero-shot baseline using Helsinki-NLP's MarianMT with its default tokenizer against a fine-tuned model leveraging a Tigrinya-specific tokenizer. The fine-tuned model consistently outperformed the baseline, achieving BLEU and chrF scores of 21 and 19.50 for English-to-Tigrinya, respectively, and a chrF score of 16.20 in the reverse direction.

In contrast, the baseline model, which employs only pretrained MarianMT weights and a generic tokenizer, yielded substantially lower performance BLEU scores of 19 (English to Tigrinya) and 17 (Tigrinya to English), with corresponding chrF scores of 10.49 and 9.39. This performance gap highlights the inherent limitations of out-of-the-box multilingual models when applied to morphologically rich, low-resource languages such as Tigrinya.

Moreover, when applying multilingual pretrained models to language-specific tasks, we recognize that pretrained weights do not transfer directly or optimally without appropriate adaptation. In particular, effective embedding initialization plays a critical role in enabling proper transfer and improving downstream performance, especially in languages with unique morphological and script characteristics.

Further evaluation on an in-domain benchmark dataset confirmed the robustness of the fine-tuned model, showing consistent gains in translation quality. By comparison, the baseline's performance degraded notably on out-of-domain data, underscoring the importance of domain adaptation and linguistically informed tokenization in enhancing model generalization for low-resource settings.

Despite these improvements, the best-performing system remains considerably below human translation standards, with human English-to-Tigrinya translation achieving a BLEU score of 89. This discrepancy underscores the significant challenges that remain in bridging the quality gap between machine and human translation for underrepresented languages.

Collectively, these results demonstrate that tailored preprocessing, including language-aware tokenization, effective embedding initialization, and domain-specific fine-tuning, are essential strategies to improve neural machine translation for Tigrinya. Such approaches effectively address the complexities posed by the language's morphology and script, thereby narrowing the performance gap in low-resource machine translation.

## VII. Conclusion and Future Work

This study highlights the effectiveness of combining language-specific tokenization with fine-tuning strate-gies for improving English–Tigrinya machine translation. While pretrained multilingual models like MarianMT offer a valuable starting point, their performance on morphologically rich, low-resource languages remains limited without tailored adaptation. Our findings emphasize that translation quality significantly improves when the tokenizer is aware of the script and morphological structure of the target language. Looking ahead, future work will explore extending translation frameworks across related Geez-script languages (e.g., Amharic and Tigre), where shared linguistic structures may enable more effective cross-lingual transfer. Additionally, the role of embedding initialization during transfer from multilingual pretrained models warrants further investigation, particularly for languages with limited direct training exposure. Such efforts will contribute to building more inclusive and robust multilingual systems for underrepresented scripts.

## Limitations

Despite the improvements achieved, several limitations remain. First, the availability of high-quality, domain-diverse parallel corpora for Tigrinya is still limited, which restricts the generalizability of the model across broader contexts. Second, while our custom tokenizer improves morphological segmentation, it does not yet fully capture dialectal variation or syntactic nuances, which are important for robust translation in real-world settings. Third, although BLEU and chrF metrics provide useful quantitative insights, they may not fully reflect semantic fidelity and fluency, especially in low-resource and morphologically rich languages. Lastly, our experiments are constrained by the pretrained models' exposure to Tigrinya during multilingual training; thus, the effectiveness of transfer heavily depends on the quality of embedding initialization and remains an open research area.

## Ethical Considerations

This work recognizes the ethical challenges involved in developing NLP systems for low-resource languages like Tigrinya. All datasets were sourced from publicly available or licensed corpora, with attention to copyright and community norms. Given the linguistic diversity and sociocultural sensitivity of Tigrinya, care was taken to avoid dialectal misrepresentation and linguistic bias.

Pretrained multilingual models may carry over biases from imbalanced training data. To mitigate this, we used language-specific tokenization and human-aligned benchmarks. However, models trained on unfiltered data, especially from social media, risk generating toxic or biased outputs issues that must be addressed seriously.

While this study does not introduce new ethical risks beyond those known in multilingual NLP, it underscores the importance of involving native speaker communities in model validation and deployment. Future work will prioritize participatory data collection and inclusive evaluation to ensure responsible and culturally informed technology development.

## Acknowledgments

This research was supported by the German Academic Exchange Service (DAAD) through the Hilde Domin Programme (funding no. 57615863).

Tigrinya (TI) & English (EN)

---

ነባሪ ኣየር ኣብ ሓደ ከባቢ ዝውቱር ዝኾነ ኩነታት ኣየር እዩ። & Climate is the long-lasting weather of a particular area.

እዚ ምስ ስነ ምድራዊ ኣቀማምጣ ከባቢ ቀጥታዊ ርክብ ኣለዎ። & It has a direct connection with the geographical characteristics of a region.

ደጉዓ ዝኾኑ ቦታታት ካብ ፀፍሒ ባሕሪ ንላዕሊ ካብ 2,500–4,000 ሜተር ዝኸውን ብራኸ ኣለዎም። & Highland regions are located from 2,500 to 4,000 meters above sea level.

እዚ ነባሪ ኣየር ከም ሩዝ፣ ዓይኒዓተርን ዓተርን ዝበሉ ዘራእቲን ከም ሰሰግን ኣወሱዳን ዝበሉ ቅመማትን ንምፍራይ ምቹው እዩ። & This climate is ideal for growing crops like rice, wheat, peas, chickpeas, and spices such as basil and black seed.

Table IV
Sample Evaluation Dataset Snippet from the Educational Domain with Sentence-level Alignment